\documentclass[sigconf]{acmart}

\usepackage{booktabs} 

\usepackage{graphicx}
\usepackage{subfig}
\graphicspath{ {images/} }
\usepackage{amsmath,amssymb} 
\usepackage{color}
\usepackage{verbatim}
\usepackage{algorithm}
\usepackage{algorithmic}
\usepackage{bm}
\usepackage{multirow}

\usepackage[normalem]{ulem}
\newcommand{\rev}[1]{{\color{black}{#1}}}
\newcommand{\del}[1]{\sout{#1}}  

\setcopyright{rightsretained}

\acmDOI{10.475/123_4}

\acmISBN{123-4567-24-567/08/06}

\acmConference[SEOUL'18]{ACM Seoul conference}{October 2018}{Seoul, Korea}
\acmYear{2018}
\copyrightyear{2018}

\acmArticle{4}
\acmPrice{15.00}

\begin{document}
\title{Crossing-Domain Generative Adversarial Networks for Unsupervised Multi-Domain Image-to-Image Translation}

\author{Xuewen Yang}
\orcid{1234-5678-9012}
\affiliation{%
  \institution{Stony Brook University}
}
\email{xuewen.yang@stonybrook.edu}

\author{Dongliang Xie}
\authornote{Dr.~Xie is the contact author.}
\affiliation{%
  \institution{Beijing University of Posts and Telecommunications}
}
\email{xiedl@bupt.edu.cn}

\author{Xin Wang}
\affiliation{%
  \institution{Stony Brook University}
}
\email{x.wang@stonybrook.edu}

\fancyhead{}

\begin{abstract}
State-of-the-art techniques in Generative Adversarial Networks (GANs) have shown remarkable success in image-to-image translation from peer domain \textit{X} to domain \textit{Y} using paired image data. However, obtaining abundant paired data is a non-trivial and expensive process in the majority of applications. When there is a need to translate images across $n$ domains, if the training is performed between every two domains, the complexity of the training will increase quadratically. Moreover, training  with data from two domains only at a time cannot benefit from data of other domains, which prevents the extraction of more useful features and hinders the progress of this research area. In this work, we propose a general framework for unsupervised image-to-image translation across multiple domains, which can translate images from domain $X$ to any a domain without requiring direct training between the two domains involved in image translation. A byproduct of the framework is the reduction of computing time and computing resources, since it needs less time than training the domains in pairs as is done in state-of-the-art works. Our proposed framework consists of a pair of encoders along with a pair of GANs which learns high-level features across different domains to generate diverse and realistic samples from. Our framework shows competing results on many image-to-image tasks compared with state-of-the-art techniques. 
\end{abstract} 

%
%



\maketitle

\section{Introduction}

In this work, we define multi-domain as multiple datasets or several subsets of one dataset that are applied to complete the same task, but these datasets (or subsets) have different statistical biases. As some examples, images taken at Alps in the summer  and in the winter are considered as two different domains, while faces with hair and faces with eyeglasses form another two different domains. Under this domain definition, for faces with black hair and faces with yellow hair, the black hair and yellow hair are two different \textit{attributes} of the same domain. In \textit{multi-domain learning}, each sample $\boldsymbol{x}$ is drawn from a domain $d$ specific distribution $\boldsymbol{x}\sim p_{d}(\boldsymbol{x})$ and has a label $y\in \{0,1\}$, with $y=1$ signifying $\boldsymbol{x}$ from domain $d$, $y=0$ signifying $\boldsymbol{x}$ not from domain $d$.

Image-to-image translation is the task of learning to map images from one domain to another, e.g., mapping grayscale images to color images \cite{DBLP:conf/pkdd/CaoZZY17}, mapping images of low resolution to images of high resolution \cite{DBLP:conf/cvpr/LedigTHCCAATTWS17}, changing the seasons of scenery images \cite{8237506}, and reconstructing photos from edge maps \cite{Isola2017ImagetoImageTW}. The most significant improvement in this research field came with the application of Generative Adversarial Networks (GANs) \cite{NIPS2014_5423,NIPS2016_6544}.

\begin{figure}[!t]
\begin{minipage}{1\linewidth}
\centering
\subfloat[]{\label{m:a}\includegraphics[scale=.19]{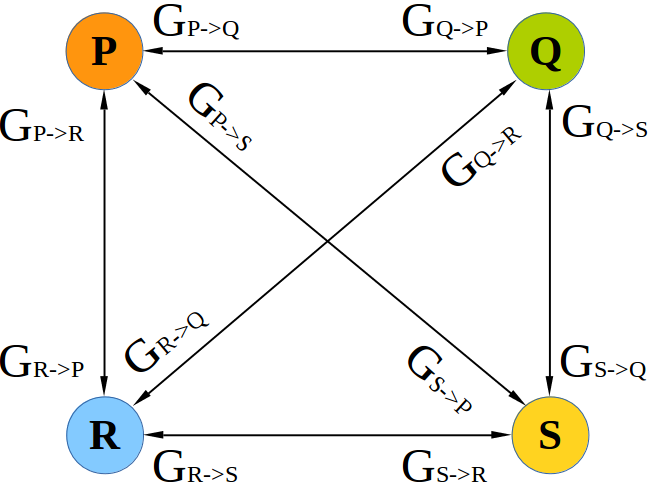}}
\subfloat[]{\label{m:b}\includegraphics[scale=.19]{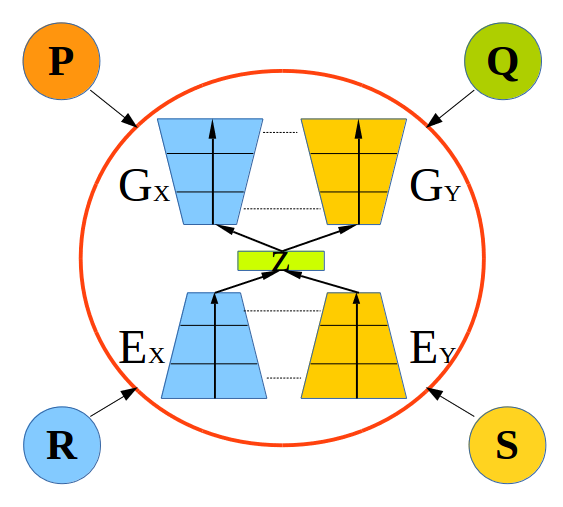}}
\end{minipage}\par
\caption{Image-to-image translation of 4 domains. (a) CycleGAN needs $4\times 3$ generators. (b) Our model only needs 2 encoder-generator pairs. In every iteration, we randomly pick two domains, and sample two batches of training data from the domains to train the model. The two encoders first encode domain information into a latent code $\boldsymbol{z}$ using two encoders $E_X$ and $E_Y$ and then generate two samples of the two domains using the generators $G_X$ and $G_Y$.}
\label{different_models}
\vspace{-0.2in}
\end{figure}

The image-to-image translation can be performed in supervised \cite{Isola2017ImagetoImageTW} or unsupervised way \cite{8237506}, with the unsupervised one becoming more popular since it does not need to collect ground-truth pairs of samples. Despite the quick progress of research on image-to-image translation, state-of-the-art results for unsupervised translation are still not satisfactory. In addition, existing research generally focuses on image-to-image translation between two domains, which is limited by two drawbacks. First, the translation task is specific to two domains, and the model has to be retrained when there is a need to perform image translation between another pair of similar domains. Second, it can not benefit from the features of multiple domains to improve the training quality. We take the most representative work in this research field CycleGAN \cite{8237506} as an example to illustrate the first limitation. The translation between two image domains $X$ and $Y$ is achieved with two generators, $G_{X\rightarrow Y}$ and $G_{Y\rightarrow X}$. However, this model is inefficient in completing the task of multi-domain image translation. To derive mappings across all $n$ domains, it has to train $n(n-1)$ generators, as shown in Fig.~\ref{m:a}.

To enable more efficient multi-domain image translation with unsupervised learning where image pairing across domains is not predefined, we propose Crossing-Domain GAN (CD-GAN), which is a multi-domain encoding generative adversarial network that consists of a pair of encoders and a pair of generative adversarial networks (GANs). We would like the encoders to efficiently encode the information of all domains to form a high-level feature space with an encoding process, then images of different domains will be translated by decoding the high-level features with a decoding process. CD-GAN achieves this goal with the integrated use of three techniques. First, the two encoders are constrained by a \textit{weight sharing} scheme, where the two encoders (or the two generators) share the same weights at both the highest-level layers and the lowest-level layers.
This ensures that the two encoders can encode common high-level semantics as well as low-level details to obtain the feature space, based on which generators can decode the high-level semantics and low-level details correctly to generate images of different domains. 
Second, we use a selected or existing label to guide the generator to generate images of a corresponding domain from the high-level features learnt. Third, we propose an efficient training algorithm that jointly train the model across domains by randomly selecting two domains to train at each iteration.

Different from ~\cite{NIPS2016_6544} where only weights at high-level layers of generators are shared, in CD-GAN, we propose the concurrent sharing of the lowest-level and the highest-level layers at both the encoders and the generators to improve the quality of image translation between {\em any two} domains. The sharing of highest layers between two encoders helps to enable more flexible cross-domain image translation, while the sharing of the lowest layers across domains helps improve the training quality by taking advantage of the transferring learning across domains.

The contributions of our work are as follows:
\begin{itemize}
\item We propose CD-GAN that learns mappings across multiple domains using only two encoder-generator pairs.
\item We propose the concurrent use of weight-sharing at highest-level and lowest-level layers of both encoders and generators to ensure that CD-GAN generates images with sufficient useful high-level semantics and low-level details across all domains.
\item We leverage domain labels to make a conditional GAN training that greatly improves the performance of the model.
\item We introduce a cross-domain training algorithm that efficiently and sufficiently trains the model by randomly taking samples from two of domains at a time. CD-GAN can fully exploit data from all domains to improve the training quality for each individual domain.
\end{itemize}

Our experiment results demonstrate that when trained on more than two domains, our method achieves the same quality of image translation between any two domains as compared to directly training for translation between the pair. However, our model is established with much less training time and can generate better quality images for a given amount of time. We also show how CD-GAN can be successfully  applied  to a variety of unsupervised multi-domain image-to-image translation problems.

The remainder of this paper is organized as follows. Section~\ref{relatedwork} reviews the relevant research for image-to-image translation problems. Section~\ref{mod} describes our model and training method in details. Section~\ref{exp} presents our evaluation metrics, experimental methodology, and the evaluation results of the model's accuracy and efficiency on different datasets. Finally, we discuss some limitations of our work and conclude our work in Section~\ref{conclusion}.

\section{Related Work}
\label{relatedwork}

\subsection{Generative Adversarial Networks (GANs)}

GANs  \cite{NIPS2014_5423} were introduced to model a data distribution using independent  latent variables. Let $\boldsymbol{x} \sim p(\boldsymbol{x})$ be a random variable representing the observed data and $\boldsymbol{z} \sim p(\boldsymbol{z})$ be a latent variable. The observed variable is assumed to be generated by the latent variable, i.e., $\boldsymbol{x} \sim p_{\boldsymbol{\theta}}(\boldsymbol{x} \vert \boldsymbol{z})$, where $p_{\boldsymbol{\theta}}(\boldsymbol{x} \vert \boldsymbol{z})$ can be explicitly represented by a generator in GANs. GANs are built on top of neural networks, and can be trained with gradient descent based algorithms.

The GAN model is composed of a discriminator $D_{\boldsymbol{\phi}}$, along with the generator $G_{\boldsymbol{\theta}}$. The training involves a min-max game between the two networks. The discriminator $D_{\boldsymbol{\phi}}$ is trained to differentiate `fake' samples generated from the generator $G_{\boldsymbol{\theta}}$ from the `real' samples from the true data distribution $p(\boldsymbol{x})$. The generator is trained to synthesize samples that can fool the discriminator by mistaking the generated samples for genuine ones. They both can be implemented using neural networks.

At the training phase, the discriminator parameters $\boldsymbol{\phi}$  are firstly updated, followed by the update of the generator parameters $\boldsymbol{\theta}$. The objective function is given by:

\begin{equation}
\begin{aligned}
\min_{\boldsymbol{\theta}}\max_{\boldsymbol{\phi}}V(D,G)=&\mathbb{E}_{x\sim p(x)}[\log D_{\boldsymbol{\phi}}(x)] \\
& + \mathbb{E}_{\boldsymbol{z}\sim p(\boldsymbol{z})}[\log (1-D_{\boldsymbol{\phi}}(G_{\boldsymbol{\theta}}(\boldsymbol{z})))]
\end{aligned}
\end{equation}

The samples can be generated by sampling $\boldsymbol{z}\sim p(\boldsymbol{z})$, then $\hat{\boldsymbol{x}}=G_{\boldsymbol{\theta}}(\boldsymbol{z})$, where $p(\boldsymbol{z})$ is a prior distribution, for example, a multivariate Gaussian.

\subsection{Image-To-Image Translation}

\textit{Image-to-image translation} problem is a kind of image generation task that given an input image $\boldsymbol{x}$ of domain \textit{X}, the model maps it into a corresponding output image $\boldsymbol{y}$ of another domain \textit{Y}. It learns a mapping between two domains given sufficient training data \cite{Isola2017ImagetoImageTW}. Early works on image-to-image translation mainly focused on tasks where the training data of domain $X$ are similar to the data of domain $Y$ \cite{Gupta:2012:ICU:2393347.2393402,Liu:2008:IC:1409060.1409105}, and the results were often unrealistic and not diverse.

In recent years, deep generative models have shown increasing capability of synthesizing diverse, realistic images that capture both fine-grained details and global coherence of natural images \cite{2015arXiv150204623G,2015arXiv151106434R,Kingma2014}.
With Generative Adversarial Networks (GANs) \cite{Isola2017ImagetoImageTW,8237506,pmlrv70kim17a}, recent studies have already taken significant steps in image-to-image translation. In \cite{Isola2017ImagetoImageTW}, the authors use a conditional GAN on different image-to-image translation tasks, such as synthesizing photos from label maps and reconstructing objects from edge maps. However, this method requires input-output image pairs for training, which is in general not available in image-to-image translation problems. For situations where such training pairs are not given, in \cite{8237506}, the authors proposed CycleGAN to tackle unsupervised image-to-image translation. With a pair of Generators $G$ and $F$, the model not only learns a mapping $G: X\rightarrow Y$ using an adversarial loss, but constrains this mapping with an inverse mapping $F: Y\rightarrow X$. It also introduces a cycle consistency loss to enforce $F(G(X)) \approx X$, and vice versa. In  settings where paired training data are not available, the authors showed promising qualitative results. The authors in \cite{pmlrv70kim17a} and \cite{8237572} use similar idea to solve the unsupervised image-to-image translation tasks.

These approaches only tackle the problems of translating images between two domains, and have two major drawbacks. First, when applied to $n$ domains, these approaches need $n(n-1)$ generators to complete the task, which is computationally inefficient. To train all models, it would either require a significant amount of time to complete if the training is performed on one GPU, or it will require a lot of hardware and computing resources if training is run over multiple GPUs.
Second, as each model is trained with only two datasets, the training cannot benefit from the data of other domains.

Our work is inspired by \textit{multimodal learning} \cite{Ngiam:2011:MDL:3104482.3104569}, which shows that data features can be better extracted using one modality if multiple modalities are present at feature learning time. The intuition of our method is that if we can encode the information of different domains together and generate a high-level feature space, it would be possible to decode the high-level features to build images of different domains. 
In this work, rather than generating images from random noise, we incorporate an encoding process into a GAN model. The image-to-image translation can be achieved by first encoding real images into high-level features, and then generating images of different domains using the high-level features through a decoding process. The encoding process and the decoding process are constrained by a weight-sharing technique that both the highest layer and the lowest layer are shared across the two encoders as well as the two generators. Sharing the high-level layers makes sure that the generated images are semantically correct, while sharing the low-level layers ensures that important low-level features be captured and transferred between domains. Our model is trained end-to-end using data from all $n$ domains.

\section{Cross-Domain Generative Adversarial Network}
\label{mod}

To conduct unsupervised multi-domain image-to-image translation, a direct approach is to train a CycleGAN for every two domains. While this approach is straightforward, it is inefficient as the number of  training models increases quadratically with the number of domains. If we have $n$ domains, we have to train $n(n-1)$ generators, as shown in Fig.~\ref{m:a}. In addition, since each model only utilizes data from two domains \del{$X$ and $Y$} to train, the training cannot benefit from the useful features of other domains.

To tackle these two problems, a possible way is to encode useful information of all domains into common high level features, and then to decode the high-level features into images of different domains. Inspired by work \cite{Ngiam:2011:MDL:3104482.3104569} from \textit{mutimodal learning}, where training data are from multiple modalities, we propose to build a multi-domain image translation model that can encode information of multiple domains into a set $Z$ of high-level features, and then use features in $Z$ to reconstruct data of different domains or to do image-to-image translation. The overview of the model applied to 4 domains is shown in Fig.~\ref{m:b}, where only one model is used.

In this section, we first present our proposed CD-GAN model, then describe how  image translation can be performed across domains, and finally introduce our cross-domain training method. 





\subsection{CD-GAN with Double Layer Sharing}

We first describe how to apply our model to multi-domain image-to-image translation in general then illustrate it using two domains as an example.
As shown in Fig.~\ref{mm:b}, our proposed CD-GAN model consists of a pair of encoders followed by a pair of  GANs. Taking domain $X$ and $Y$ as an example, the two encoders $E_X$ and $E_Y$ encode domain information from $X$ and $Y$ into a set of high-level features contained in a set $Z$. Then from a high-level feature $z$ in space $Z$,
we can generate images that fall into domain $X$ or $Y$. The generated images are then evaluated by the corresponding discriminators $D_X$ and $D_Y$ to see whether they look real and cannot be identified as generated ones. For example, following the red arrows, the input image $\boldsymbol{x}$ is first encoded into a high-level feature $\boldsymbol{z}_x$, then $\boldsymbol{z}_x$ is decoded to generate the image $\hat{\boldsymbol{y}}$. The image $\hat{\boldsymbol{y}}$ is the translated image in domain $Y$. Similar processes exist for image $\boldsymbol{y}$.


Our model is also constrained by a reconstruction process shown in Fig.~\ref{mm:c}. For example, following the red arrows, the input image $\boldsymbol{x}$ is first encoded into a high-level feature $\boldsymbol{z}_x$, then $\boldsymbol{z}_x$ is decoded to generate the image $\boldsymbol{x}\prime$, which is a reconstruction of the input image. Similar processes exist for image $\boldsymbol{y}$.

Learning with deep neural networks involves hierarchical feature representation.  In order to support flexible cross-domain image translation and also to improve the training quality, we propose the use of {\em double-layer sharing} where the highest-level and the lowest-level layers of the two encoders share the same weights and so does the two generators. By enforcing the layers that decode high-level features in GANs to share weights, the images generated by different generators can have some common high-level semantics. The layers that decode low-level details then map the high-level features to images in individual domains.

Sharing weights of low-level layers has the benefit of transferring low-level features of one domain to the other, thus making the image-to-image translation more close to real images in the respective domains. Besides, sharing layers reduces the complexity of the model, making it more resistant to the over-fitting problem.


\subsection{Conditional Image Generation}
\label{cig}
In state-of-the-art techniques, like CycleGAN, each domain is described by a specific generator, thus there is no need 
to inform the generator which domain the input image is generated to. However, in our model, multiple domains share two generators. For an input image, we have to include an auxiliary variable to guide the generation of image for a specific domain. The only information we have is the domain labels. To make use of this information, the inputs of the model are not images $\boldsymbol{x}$, $\boldsymbol{y}$, but image pairs $(\boldsymbol{x}, \boldsymbol{l}_y)$ and $(\boldsymbol{y}, \boldsymbol{l}_x)$ where the labels $\boldsymbol{l}_y$ and $\boldsymbol{l}_x$ inform the generators which domains to generate an image for. These image pairs are not the same as the image pairs of supervised image-to-image generation tasks, which are $(\boldsymbol{x}, \boldsymbol{y})$. Thus no matter which domain images are the input, the model can always generate images of a domain of interest.


\begin{figure}[!t]
\begin{minipage}{1\linewidth}
\centering
\subfloat[]{\label{mm:b}\includegraphics[scale=.2]{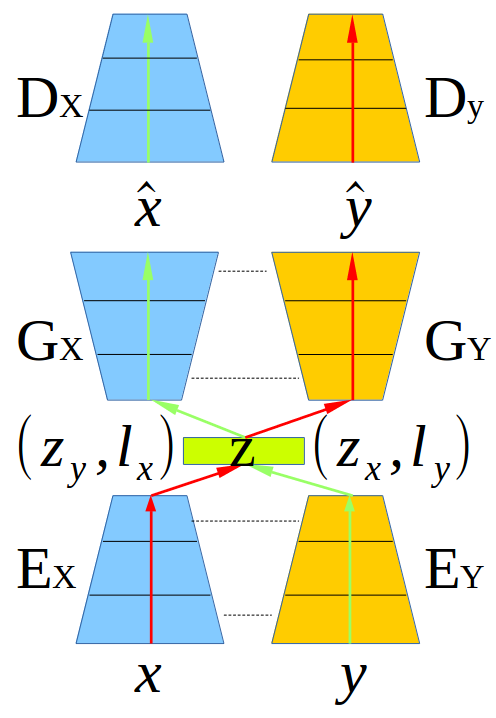}}
\subfloat[]{\label{mm:c}\includegraphics[scale=.2]{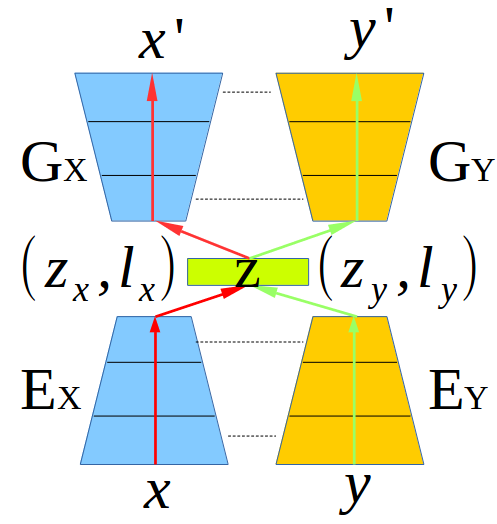}}
\end{minipage}\par
\caption{The proposed CD-GAN model. (a) The translation mappings: the input image $\boldsymbol{x}$ is first encoded as a latent code $\boldsymbol{z}_x$ through $E_X(\boldsymbol{x})$, which is then decoded into a translated image $\hat{\boldsymbol{y}}$ through $G_Y(\boldsymbol{z}_x, \boldsymbol{l}_y)$. The process is identified with red arrows. There is a similar process for the image $\boldsymbol{y}$. $D_X$ and $D_Y$ are adversarial discriminators for the respective domains to evaluate whether the translated images are realistic. (b) The reconstruction mappings: the input image $\boldsymbol{x}$ is first encoded as a latent code $\boldsymbol{z}_x$ through $E_X(\boldsymbol{x})$, which is then decoded into a reconstructed image $\boldsymbol{x}\prime$ through the generator $G_X(\boldsymbol{z}_x, \boldsymbol{l}_x)$. The process is signified in red arrows. A similar process exists for image $\boldsymbol{y}$. Note: the dashed lines indicate that the two layers share the same parameters.}
\label{model:loss}
\vspace{-0.2in}
\end{figure}

We denote the data distributions as $\boldsymbol{x}\sim p(\boldsymbol{x})$ and $\boldsymbol{y}\sim p(\boldsymbol{y})$. As illustrated in Fig. \ref{model:loss}, our model includes four mappings, two translation mappings $X\rightarrow Z\rightarrow Y$, $Y\rightarrow  Z\rightarrow X$ and two reconstruction mappings $X\rightarrow Z \rightarrow X$, $Y\rightarrow Z\rightarrow Y$. The translation mappings constrain the model by a GAN loss, while the reconstruction mappings constrain the model by a reconstruction loss. To further constrain the auxiliary variable, we introduce a classification loss by applying a classifier to classify the real or generated images into different domains. The intuition is that if images are generated with the guidance of the auxiliary variable, then it can be correctly classified into the domain specified by the auxiliary variable. Next, we introduce these model losses in more details as follows.

\textbf{GAN Losses} Following the translation mapping $X\rightarrow Z\rightarrow Y$, we can translate image $\boldsymbol{x}$ from domain $X$ to $\hat{\boldsymbol{y}}$ of domain $Y$ using $\boldsymbol{z}_x = E_{X}(\boldsymbol{x})$, $\hat{\boldsymbol{y}}=G_{Y}(\boldsymbol{z}_x, \boldsymbol{l}_y)$. With the purpose of improving the quality of the generated samples, we apply adversarial loss. We express the objective as:

\begin{equation}
\begin{aligned}
\mathcal{L}_{GAN_{Y}}&=\mathbb{E}_{\boldsymbol{y}\sim p(\boldsymbol{y})}\log (D_{Y}(\boldsymbol{y})) \\ &
 + \mathbb{E}_{\boldsymbol{x}\sim p(\boldsymbol{x})}\log (1-D_{Y}(G_{Y}(E_{X}(\boldsymbol{x}),\boldsymbol{l}_y)))
\end{aligned}
\end{equation}
where $G_Y$ tries to generate images $\hat{\boldsymbol{y}}=G_{Y}(\boldsymbol{z}_x,\boldsymbol{l}_y)$ that look similar to images from domain $Y$, while $D_{Y}$ aims to distinguish between translated samples $\hat{\boldsymbol{y}}$ and real samples $\boldsymbol{y}$. The similar adversarial loss for $Y\rightarrow  Z\rightarrow X$ is

\begin{equation}
\begin{aligned}
\mathcal{L}_{GAN_{X}}&=\mathbb{E}_{\boldsymbol{x}\sim p(\boldsymbol{x})}\log (D_{X}(\boldsymbol{x})) \\ &
 + \mathbb{E}_{\boldsymbol{y}\sim p(\boldsymbol{y})}\log (1-D_{X}(G_{X}(E_{Y}(\boldsymbol{y}),\boldsymbol{l}_x)))
\end{aligned}
\end{equation}

The total GAN loss is:
\begin{equation}
\mathcal{L}_{GAN}=\mathcal{L}_{GAN_{X}} + \mathcal{L}_{GAN_{Y}}
\end{equation}

\textbf{Reconstruction Loss}
The reconstruction mappings $X\rightarrow Z \rightarrow X$, $Y\rightarrow Z\rightarrow Y$ encourage the model to encode enough information to the high-level feature space $Z$ from each domain. The input can then be reconstructed by the generators. The reconstruction process of domain $X$ is $\boldsymbol{z}_x = E_{X}(\boldsymbol{x})$, $\boldsymbol{x}\prime=G_{X}(\boldsymbol{z}_x, \boldsymbol{l}_x)$. Similar reconstruction process exists for domain $Y$. With $l_2$ distance as the loss function, the reconstruction loss is:

\begin{equation}
\begin{aligned}
\mathcal{L}_{rec}&=\mathbb{E}_{\boldsymbol{x}\sim p(\boldsymbol{x})}(\vert \vert \boldsymbol{x} - G_{X}(E_{X}(\boldsymbol{x}),\boldsymbol{l}_x) \vert \vert_{2}) \\
&+ \mathbb{E}_{\boldsymbol{y}\sim p(\boldsymbol{y})}(\vert \vert \boldsymbol{y} - G_{Y}(E_{Y}(\boldsymbol{y}),\boldsymbol{l}_y) \vert \vert_{2})
\end{aligned}
\end{equation}

\textbf{Latent Consistency Loss}
With only the above losses, the encoding part is not well constrained. We constrain the encoding part using a latent consistency loss. Although $\boldsymbol{x}$ is translated to $\hat{\boldsymbol{y}}$, which is in domain $Y$, $\hat{\boldsymbol{y}}$ is still semantically similar to $\boldsymbol{x}$. Thus, in the latent space $Z$, the high-level feature of $\boldsymbol{x}$ should be close to that of $\hat{\boldsymbol{y}}$. Similarly, the high-level feature of $\boldsymbol{y}$ in domain $Y$ should be close to the  high-level feature of $\hat{\boldsymbol{x}}$ in domain $X$.  
The latent consistency loss is the following:

\begin{equation}
\begin{aligned}
\mathcal{L}_{lcl}&=\mathbb{E}_{\boldsymbol{x}\sim p(\boldsymbol{x})}(\vert \vert E_{X}(\boldsymbol{x}) - E_{Y}(G_{Y}(E_{X}(\boldsymbol{x}),\boldsymbol{l}_y)) \vert \vert) \\
&+\mathbb{E}_{\boldsymbol{y}\sim p(\boldsymbol{y})}(\vert \vert E_{Y}(\boldsymbol{y}) - E_{X}(G_{X}(E_{Y}(\boldsymbol{y}),\boldsymbol{l}_x)) \vert \vert)
\end{aligned}
\end{equation}

\textbf{Classification Loss}
We consider $n$ domains as $n$ categories in the classification problems. We use a network $C$, which is an auxiliary classifier, on top of the general discriminator $D$ to measure whether a sample (real or generated) belongs to a specific fine-grained category. The output of the classifier $C$ represents the posterior probability $P(c\vert \boldsymbol{x})$. Specifically, there are four classification losses, i.e., for real data $\boldsymbol{x}$, $\boldsymbol{y}$, and generated data $\hat{\boldsymbol{x}}$, $\hat{\boldsymbol{y}}$. For image-label pairs ($\boldsymbol{x}$, $\boldsymbol{l}_{x}$) and ($\boldsymbol{y}$, $\boldsymbol{l}_{y}$) with $\boldsymbol{l}_{x}\sim p(\boldsymbol{l}_{x})$ and $\boldsymbol{l}_{y}\sim p(\boldsymbol{l}_{y})$ our goal is to translate $\boldsymbol{x}$ to $\hat{\boldsymbol{y}}$ with label $\boldsymbol{l}_{y}$, and to translate $\boldsymbol{y}$ to $\hat{\boldsymbol{x}}$ with label $\boldsymbol{l}_{x}$. The four classification losses are:

\begin{equation}
\begin{aligned}
\mathcal{L}_{c}&=-\mathbb{E}_{\boldsymbol{x}\sim p(\boldsymbol{x}), \boldsymbol{l}_{x}\sim p(\boldsymbol{l}_{x})}[\log P(\boldsymbol{l}_{x}\vert \boldsymbol{x})] \\
&=-\mathbb{E}_{\boldsymbol{y}\sim p(\boldsymbol{y}), \boldsymbol{l}_{y}\sim p(\boldsymbol{l}_{y})}[\log P(\boldsymbol{l}_{y}\vert \boldsymbol{y})] \\
&=-\mathbb{E}_{\boldsymbol{x}\sim p(\boldsymbol{x}), \boldsymbol{l}_{y}\sim p(\boldsymbol{l}_{y})}[\log P(\boldsymbol{l}_{y}\vert G_{Y}(E_{X}(\boldsymbol{x}), \boldsymbol{l}_{y}))] \\
&=-\mathbb{E}_{\boldsymbol{y}\sim p(\boldsymbol{y}), \boldsymbol{l}_{x}\sim p(\boldsymbol{l}_{x})}[\log P(\boldsymbol{l}_{x}\vert G_{X}(E_{Y}(\boldsymbol{y}), \boldsymbol{l}_{x}))]
\end{aligned}
\end{equation}

This loss can be used to optimize discriminators $D_{X}$, $D_{Y}$, generators $G_{X}$, $G_{Y}$, and encoders $E_{X}$, $E_{Y}$.

\textbf{Cycle Consistency Loss}
Although the minimization of GAN losses ensures that $G_Y(E_{X}(\boldsymbol{x}), \boldsymbol{l}_{y})$ produce a sample $\hat{\boldsymbol{y}}$ similar to samples drawn from $Y$, the model still can be unstable and prone to failure. To tackle this problem, we further constrain our model with a cycle-consistency loss \cite{8237506}. To achieve this goal,  we want mapping from domain $X$ to domain $Y$ and then back to domain $X$ to reproduce the original sample, i.e., $G_{X}(E_{Y}(G_{Y}(E_{X}(\boldsymbol{x}), \boldsymbol{l}_{y})), \boldsymbol{l}_{x}) \approx \boldsymbol{x}$ and $G_{Y}(E_{X}(G_{X}(E_{Y}(\boldsymbol{y}), \boldsymbol{l}_{x})), \boldsymbol{l}_{y}) \approx \boldsymbol{y}$. Thus, the cycle-consistency loss is:

\begin{equation}
\begin{aligned}
\mathcal{L}_{cyc}&=\mathbb{E}_{\boldsymbol{x}\sim p(\boldsymbol{x})}[\vert \vert G_{X}(E_{Y}(G_{Y}(E_{X}(\boldsymbol{x}), \boldsymbol{l}_{y})), \boldsymbol{l}_{x}) - \boldsymbol{x} \vert \vert] \\
&+ \mathbb{E}_{\boldsymbol{y}\sim p(\boldsymbol{y})}[\vert \vert G_{Y}(E_{X}(G_{X}(E_{Y}(\boldsymbol{y}), \boldsymbol{l}_{x})), \boldsymbol{l}_{y}) - \boldsymbol{y} \vert \vert]
\end{aligned}
\end{equation}

\textbf{Final Objective of CD-GAN}
To sum up, the goal of our approach is to minimize the following objective:
\begin{equation}
\begin{aligned}
\mathcal{L}(E, G, D) & =\mathcal{L}_{GAN}  + \alpha_{0} \mathcal{L}_{rec} + \alpha_{1} \mathcal{L}_{lcl} + \alpha_{2} \mathcal{L}_{c} + \alpha_{3} \mathcal{L}_{cyc}
\end{aligned}
\end{equation}
where $E$, $G$, and $D$ signify encoders $E_X$, $E_Y$, generators $G_X$, $G_Y$, and discriminators $D_X$, $D_Y$, and $\alpha_{0}$, $\alpha_{1}$, $\alpha_{2}$, $\alpha_{3}$ control the relative importance of the losses. Same as solving a regular GAN problem, training the model involves the solving  of a min-max problem, where $E_{X}$,$E_{Y}$, $G_{X}$, and $G_{Y}$ aim to minimize the objective, while $D_{X}$ and $D_{Y}$ aim to maximize it.

\begin{equation}
E^{\ast}, G^{\ast} = arg \min_{E, G} \max_{D} \mathcal{L}(E, G, D)
\end{equation}


\subsection{Cross-Domain Training}

Our proposed model has two encoder-generator pairs, but we have data from $n$ domains. To train the model using samples of all domains equally, we introduce a cross-domain training algorithm. As shown in Fig.~\ref{m:b}, 
there are 4 domains. At each iteration, we randomly select two domains $R$ and $S$, and feed training data of these two domains into the model. At the next iteration, we might take another two domains $P$ and $Q$, and perform the same training. We train the model using all data samples of $4$ domains at every epoch for several iterations. The training algorithm is shown in Algorithm ~\ref{training}. \textit{Cross-domain training} ensures the model to learn a generic feature representation of all domains by training the model equally on independent domains.

\begin{algorithm}
\caption{Joint domain training on CD-GAN using mini-batch stochastic gradient descent}
\begin{algorithmic}
\STATE \textbf{Require:} Training samples from $n$ domains
\STATE \textbf{Initialize} $\boldsymbol{\theta}_E^X$, $\boldsymbol{\theta}_E^Y$,$\boldsymbol{\theta}_G^X$, $\boldsymbol{\theta}_G^Y$,$\boldsymbol{\theta}_D^X$, and $\boldsymbol{\theta}_D^Y$ with the shared network connection weights set to the same values.
\WHILE{Training loss has not converged}
	
	\STATE Randomly draw two domains $X$ and $Y$ from $n$ domains
	\STATE Randomly draw $N$ samples from the two domains, \{$\boldsymbol{x}_1, \boldsymbol{x}_2, \ldots \boldsymbol{x}_N; \boldsymbol{y}_1, \boldsymbol{y}_2, \ldots \boldsymbol{y}_N$\}
	\STATE Get the domain labels of the samples from the two domains, $\{ \boldsymbol{l}_X^i, \boldsymbol{l}_Y^i \}_{i=1}^{N}$
	
	\STATE \textbf{(1) Update} $\boldsymbol{D}_X, \boldsymbol{D}_Y$ \textbf{with fixed} $\boldsymbol{G}_X, \boldsymbol{G}_Y, \boldsymbol{E}_X, \boldsymbol{E}_Y$	
	
	\STATE Generate fake samples using the real ones
	
	\begin{align*}
	\hat{\boldsymbol{x}}_i = G_X(E_Y(\boldsymbol{y}_i), \boldsymbol{l}_x^i),\hat{\boldsymbol{y}}_i = G_Y(E_X(\boldsymbol{x}_i), \boldsymbol{l}_y^i), i=1 \ldots N
	\end{align*}
	
	\STATE Update $\boldsymbol{\theta}_D=(\boldsymbol{\theta}_D^X, \boldsymbol{\theta}_D^Y)$  according to the following gradients
	\begin{align*}
	& \nabla_{\theta_D} \bigg [ \frac{1}{N} \sum_{i=1}^{N}\Big [-\log D_X(\boldsymbol{x}_i) - \log(1-D_X(\hat{\boldsymbol{x}}_i)) -\log D_Y(\boldsymbol{y}_i) \\
	&- \log(1-D_Y(\hat{\boldsymbol{y}}_i)) + \alpha_2 \big[ \log P(\boldsymbol{l}_x\vert \boldsymbol{x}_i)+  \log P(\boldsymbol{l}_y\vert \boldsymbol{y}_i)\big]\Big ] \bigg ]
	\end{align*}
	
	\STATE \textbf{(2) Update} $\boldsymbol{E}_X, \boldsymbol{E}_Y, \boldsymbol{G}_X, \boldsymbol{G}_Y$ \textbf{with fixed} $\boldsymbol{D}_X, \boldsymbol{D}_Y$
	\STATE Update $\boldsymbol{\theta}_{E,G}=(\boldsymbol{\theta}_{E}^{X}, \boldsymbol{\theta}_{E}^{Y}, \boldsymbol{\theta}_{G}^{X}, \boldsymbol{\theta}_{G}^{Y})$  according to the following gradients
	\begin{align*}
	& \nabla_{\theta_{E,G}} \bigg [\frac{1}{N} \sum_{i=1}^{N}\Big[\log(1-D_X(\hat{\boldsymbol{x}}_i)) + \log(1-D_Y(\hat{\boldsymbol{y}}_i)) \\
	&+ \vert \vert \boldsymbol{x}_i - G_X(E_X(\boldsymbol{x}_i), \boldsymbol{l}_{x}^i) \vert \vert_2 + \vert \vert \boldsymbol{y}_i - G_Y(E_Y(\boldsymbol{y}_i), \boldsymbol{l}_{y}^i)\vert \vert_2 \\
	&+  \vert \vert E_X(\boldsymbol{x}_i)-E_Y(\hat{\boldsymbol{y}}_i)  \vert \vert + \vert \vert E_Y(\boldsymbol{y}_i)-E_X(\hat{\boldsymbol{x}}_i)  \vert \vert \\
	&+  \log P(\boldsymbol{l}_x\vert \hat{\boldsymbol{x}}_i)+  \log P(\boldsymbol{l}_y\vert \hat{\boldsymbol{y}}_i) \\
	&+ \alpha \big[\vert \vert \boldsymbol{x}_i-G_X(E_Y(\hat{\boldsymbol{y}}_i),\boldsymbol{l}_{x}^i)  \vert \vert + \vert \vert \boldsymbol{y}_i-G_Y(E_X(\hat{\boldsymbol{x}}_i),\boldsymbol{l}_{y}^i)  \vert \vert\big] \Big] \bigg ]
	\end{align*}

\ENDWHILE
\end{algorithmic}
\label{training}
\end{algorithm}

\section{experiment}
\label{exp}

In this section, we conduct experiments over three datasets to compare our proposed model with reference models in terms of image translation quality and efficiency.

\subsection{Datasets}

To evaluate the scalability and effectiveness of our model, we test it on a variety of multi-domain image-to-image translation tasks using the following datasets:

\textbf{Alps Seasons dataset} \cite{2017arXiv171206909A} is collected from images on Flickr. The images are categorized into four seasons based on the provided timestamp of when it was taken. It consists of four categories: \textit{Spring}, \textit{Summer}, \textit{Fall}, and \textit{Winter}. The training data consists of 6053 images of four seasons, while the test data consists of 400 images.

\textbf{Painters dataset} \cite{8237506} includes painting images of four artists \textit{Monet}, \textit{Van Gogh}, \textit{Cezanne}, and \textit{Ukiyo-e}. We use 2851 images as the training set, and 200 images as the test set.

\textbf{CelebA dataset} \cite{Liu:2015:DLF:2919332.2920139} contains ten thousand identities, each of which has twenty images, i.e., two hundred thousand images in total. Each image in CelebA is annotated with 40 face attributes. We resize the initial $178 \times 218$ size images to $256 \times 256$. We randomly select 4000 images as test set and use all remaining images for training data.

We run all the experiments on a Ubuntu system using an Intel i7-6850K, along with a single NVIDIA GTX 1080Ti GPU.

\subsection{Reference Models}


We compare the performance of our proposed CD-GAN with that of two reference models:

\textbf{CycleGAN} \cite{8237506} This method trains two generators $G: X\rightarrow Y$ and $F: Y\rightarrow X$ in parallel. It not only applies a standard GAN loss respectively for $X$ and $Y$, but applies forward and backward cycle consistency losses which ensure that an image $\boldsymbol{x} $ from domain $X$ be translated to an image of domain $Y$, which can then be translated back to the domain $X$, and vice versa.

\textbf{DualGAN} \cite{8237572} This method uses a dual-GAN mechanism, which consists of a primal GAN and a dual GAN. The primal GAN learns to translate images from domain $X$ to domain $Y$, while the dual-GAN learns to invert the task. Images from either domain can be translated and then reconstructed. Thus a reconstruction loss can be used to train the model.

\rev{\textbf{UNIT} \cite{NIPS2017_6672} This method consists of two VAE-GANs with a fully shared latent space. To complete the task of image-to-image translation between $n$ domains, it needs to be trained $\frac{n\times{(n-1)}}{2}$ times.}

\rev{\textbf{DB} \cite{2017arXiv171202050H} This method addresses the multi-domain image-to-image translation problem by introducing $n$ domain-specific encoders/decoders to learn an universal shared-latent space.}

\subsection{Evaluation Metrics}

There is a challenge to evaluate the quality of synthesized images~\cite{NIPS2016_6125}. Recent works have tried using pre-trained semantic classifiers to measure the realism and discriminability of the generated images. The idea is that if the generated images look to be more close to real ones,
 classifiers trained on the real images will be able to classify the synthesized images correctly as well. Following \cite{2016arXiv160308511Z, Isola2017ImagetoImageTW, 2016arXiv160305631W}, to evaluate the  performance of the models in classifying generated images quantitatively,  we apply the metric \textit{classification accuracy}. For each experiment, we generate enough number of images of different domains, then we use a pre-trained classifier which is trained on the training dataset to classify them to different domains and calculate the classification accuracy.

\subsection{Network Architecture and Implementation}

\begin{figure}[!t]
\centering
\includegraphics[width=0.45\textwidth]{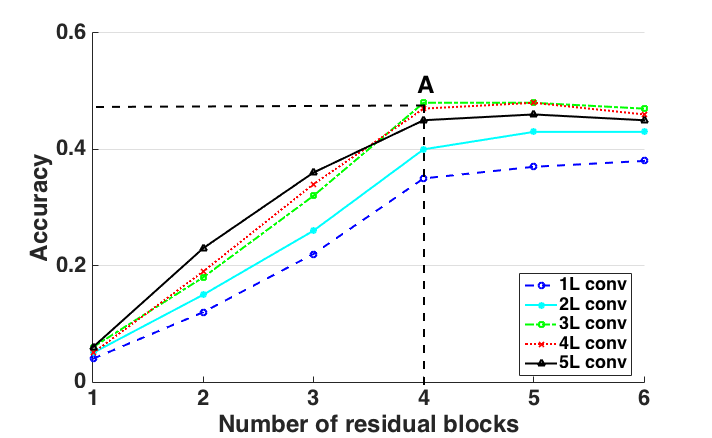}
\caption{The accuracy on varying number of residual blocks and number of convolutional layers.}
\label{res_conv}
\end{figure}

The design of the architecture is always a difficult task \cite{2015arXiv151106434R}. To get a proper model architecture, we adopt the architecture of the discriminator from \cite{Isola2017ImagetoImageTW} which has been proven to be proficient in most image-to-image generation tasks. It has 6 convolutional layers. We keep the discriminator architecture fixed and vary the architectures of the encoders and generators. Following the design of the architectures of the generators in \cite{Isola2017ImagetoImageTW}, we use two types of layers, the regular convolutional layers and the basic residual blocks \cite{DBLP:conf/cvpr/HeZRS16}. Since the encoding process is the inverse of the decoding process, we use the same layers for them but put the layers in the inverse orders. The only difference is the first layer of the encoder and the last layer of the generator. We apply $64$ channels (corresponding to different filters) for the first layer of the encoders, but $3$ channels for the last layer of the generators since the output images have only $3$ RGB channels.
 We gradually change the number of convolutional layers and the number of residual blocks until we get a satisfying architecture. We don't apply \textit{weight sharing} initially. The performance of different architectures is evaluated on the \textit{Painters} dataset and shown in Fig.~\ref{res_conv}. We can see that when the model has 3 regular convolutional layers and 4 basic residual blocks, the model has the best performance. We keep this architecture fixed for other datasets.

We then vary the number of weight-sharing layers in the encoders and the generators. We change the number of weight-sharing layers from 1 to 4. Sharing 1 layer means sharing the highest layer and the lowest layers in the encoder pair. Sharing 2 layers means sharing the highest and lowest two layers. The same sharing method applies for the generator pair (not including the output layer). 
The results are shown in table \ref{tab:share}. We found that sharing 1 layer is enough to have a good performance.

\begin{table}[!t]
\caption{Classification accuracy on number of shared layers in encoders and generators.}
\centering
\begin{tabular}{ccl}
\toprule
\textbf{\# of shared layers} & \textbf{acc. \% (Painters)} & \textbf{acc. \% (Alps Seasons)} \\
\midrule
0 & 49.75 & 29.95  \\
1 & 52.54 & 33.78 \\
2 & 52.81 & 33.54 \\
3 & 51.13 & 33.06 \\
\bottomrule
\end{tabular}
\label{tab:share}
\vspace{-0.2in}
\end{table}

In summary, for the testbed evaluation, we use two encoders each consisting of 3 convolutional layers and 4 basic residual blocks. The generators are composed with 4 basic residual blocks and 3 fractional-strided convolutional layers. The discriminators consist of a stack of 6 convolutional layers. We use LeakyReLU for nonlinearity. The two encoders share the same parameters on their layers 1 and 7, while the two generators share the same parameters on layers 1 and 6,  which is the lowest-level layer before the output layer. The details of the networks are given in table \ref{table:architecture}. We evaluate various network architectures in the evaluation parts. We fix the network architecture as in Table~\ref{table:architecture}.

\begin{table}[!t]
\caption{Network architecture for the multi-modal unsupervised image-to-image translation experiments. $cxkysz$ denote a Convolution-InstanceNorm-ReLU layer with $x$ filters, kernel size $y$, and stride $z$. $Rm$ denotes a residual block that contains two $3 \times 3$ convolutional layers with the same number of filters on both layers. $un$ denotes a $3 \times 3$ fractional-strided-Convolution-InstanceNorm-ReLU layer with $n$ filters, and stride $\frac{1}{2}$. $n_d$ denotes number of domains. $Y$ and $N$ denote whether the layer is shared or not.}
\small
\centering
\begin{tabular}{cccl}
\toprule
\textbf{Layer} & \textbf{Encoders} & \textbf{Generators} & \textbf{Discriminators}  \\
\midrule
1   &  $c64k7s1 (Y)$ & $R256 (Y)$ & $c64k3s2 (N)$
\\
2 & $c128k3s2(N)$ & $R256 (N)$ & $c128k3s2 (N)$
\\
3  & $c256k3s2 (N)$ & $R256 (N)$ & $c256k3s2 (N)$
\\
4 & $R256 (N)$ & $R256 (N)$ &  $c512k3s2 (N)$
\\
5 & $R256 (N)$ & $u256 (N)$ &  $c1024k3s2 (N)$
\\
6 & $R256 (N)$ & $u128 (Y)$& $c(1+n_d)k2s1 (N)$
\\
7 & $R256 (Y)$& $u3 (N)$ &
\\
\bottomrule
\end{tabular}

\label{table:architecture}
\vspace{-0.2in}
\end{table}

We use ADAM \cite{2014arXiv1412.6980K} for training, where the training rate is set to 0.0001 and momentums are set to 0.5 and 0.999. Each mini-batch consists of one image from domain $X$ and one image from domain $Y$. Our model has several hyper-parameters. The default values are $\alpha_{0}=10$, $\alpha_1=0.1$, ${\alpha}_2=0.1$, and ${\alpha}_3=10$. The hyper-parameters of the baselines are set to the suggested values by the authors. 

\subsection{Quantitative Results}
We evaluate our model on different datasets and compare it with baseline models.
\subsubsection{Comparison on Painters Dataset}

\begin{figure}[!t]
\centering
\includegraphics[width=0.45\textwidth]{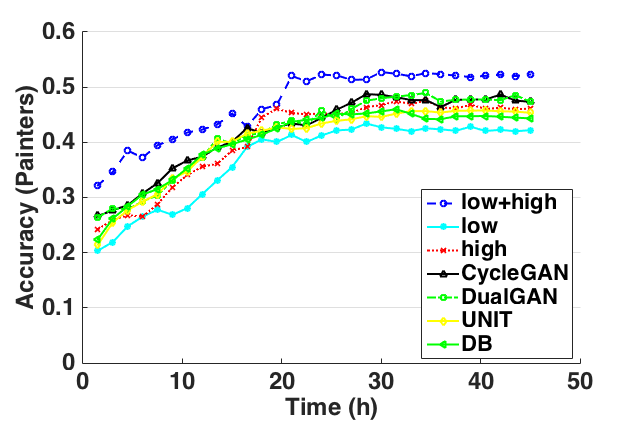}
\caption{The classification accuracy on \textit{Painters} dataset. The 7 models are the proposed model with \textit{the lowest and the highest layer sharing}, \textit{the lowest layer sharing only}, \textit{the highest layer sharing only}, CycleGAN, DualGAN, UNIT, and DB.}
\label{acc_p}
\end{figure}

To compare the proposed model with baseline models \textit{Painters} dataset, we first train the state-of-the-art VGG-11 model \cite{2014arXiv1409.1556S} on training data and get a classifier of accuracy 94.5\%. We then score synthesized images by the classification accuracy against the domain labels these photos were synthesized from. We generate around 4000 images for every 5 hours and the classification accuracies are shown in Fig.~\ref{acc_p}.

We can see that our model achieves the highest classification accuracy of 52.5\% when using both the highest layer and lowest layer sharing, with the training time less than the other reference models in reaching the peak.

\subsubsection{Comparison on Alps Seasons Dataset}
\begin{figure}[!t]
\centering
\includegraphics[width=0.45\textwidth]{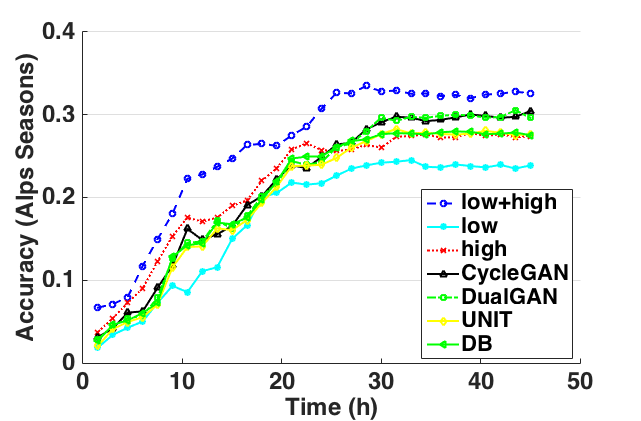}
\caption{The classification accuracy on \textit{Alps Seasons} dataset. The 7 models are the prosed model with \textit{lowest and highest level layers sharing}, \textit{lowest level layers sharing}, \textit{highest level layers sharing}, CycleGAN, DualGAN, UNIT, and DB.}
\label{acc_a}
\end{figure}

We train VGG-11 model on training data of \textit{Alps Seasons} dataset and get a classifier of accuracy 85.5\% trained on the training data. We then classify the generated images by our model and the classification accuracies are shown in Fig.~\ref{acc_a}.

Similar to Fig.~\ref{acc_p}, our model achieves the highest classification accuracy of 33.8\% with the training time less than the baseline models in reaching the peak.




\subsection{Analysis of the loss function}

We compare the ablations of our full loss. As GAN loss and cycle consistency loss are critical for the training of unsupervised image-to-image translation, we keep these two losses as the baseline model and do the ablation experiments to see the importance of other losses.

\begin{table}
  \caption{Ablation study: classification accuracy of \textit{Painters} and \textit{Alps Seasons} datasets for different losses. The following abbreviations are used: R:reconstruction loss, LCL: latent consistency loss, C: classification loss.}
  \label{tab:abl}
  \begin{tabular}{ccl}
    \toprule
    \textbf{Loss}&\textbf{acc.\% (Painters)}&\textbf{acc. \% (Alps Seasons)}\\
    \midrule
    Baseline & 35.23& 20.81\\
    Baseline + R & 36.86& 21.59\\
    Baseline + LCL & 44.42  & 25.05\\
    Baseline + C & 43.63 & 24.01\\
    Baseline + R + LCL & 45.79 & 27.19 \\
    Baseline + R + C & 44.82 & 26.63 \\
    Baseline + LCL + C & 50.74 & 32.51 \\
    Baseline + R + LCL + C & 52.54 & 33.78 \\
  \bottomrule
\end{tabular}
\end{table}

As shown in Tabel~\ref{tab:abl}, the reconstruction loss $R$  is least important with accuracy improvement of about 4.6\% on \textit{Painters} dataset and 3.7\% on \textit{Alps Seasons} dataset. The latent consistency loss $LCL$ brings the model an accuracy improvement of 26.1\% on \textit{Painters} dataset and 20.4\% on \textit{Alps Seasons} dataset. The accuracy is improved by 23.8\% on \textit{Painters} dataset and 15.4\% on \textit{Alps Seasons} dataset by the classification loss $C$.

\subsection{Qualitative Results}
We demonstrate our model on three unsupervised multi-domain image-to-image translation tasks.

\begin{figure}
\includegraphics[height=2.5in, width=2.5in]{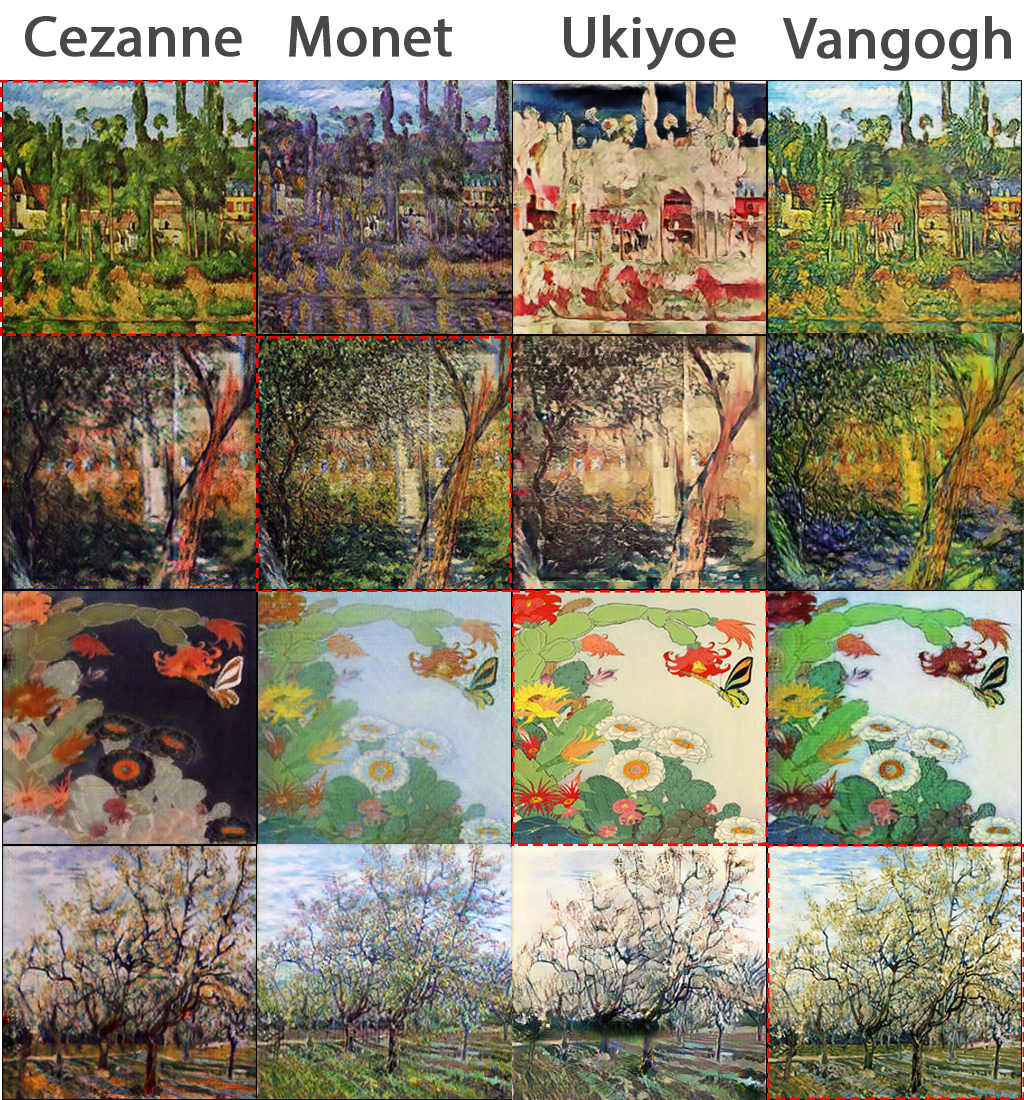}
\caption{\textit{Painters} translation results. The original images are displayed with a dashed square around. The other images are generated according to different painters.}
\label{pr}
\end{figure}

\begin{figure}
\includegraphics[height=3.2in, width=3.2in]{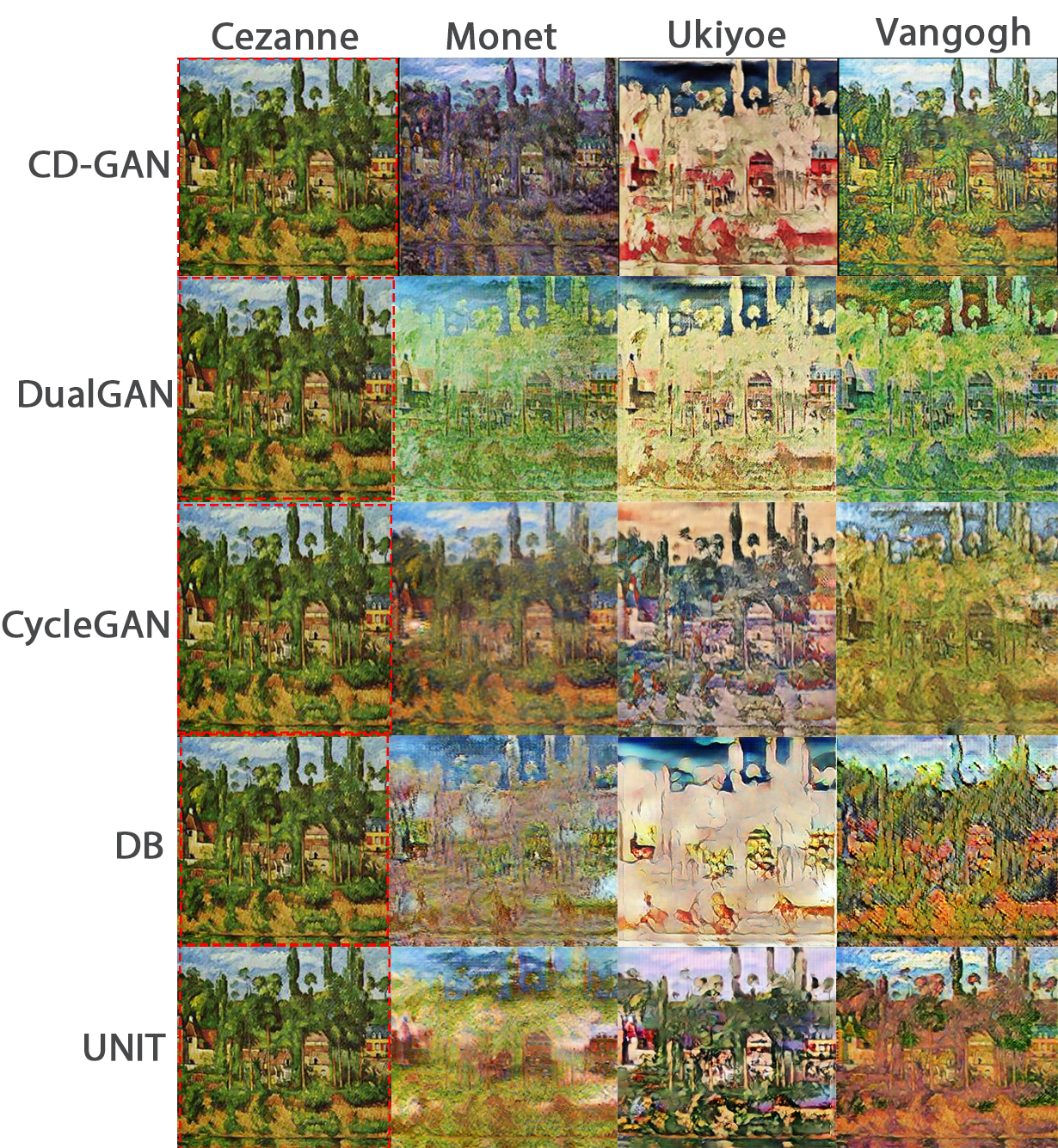}
\caption{\textit{Painters} translation results. The original images are displayed with a dashed square around. The other images are generated according to different painters.}
\label{pr_r}
\end{figure}

\textbf{Painting style transfer (Fig.~\ref{pr})} We train our model on \textit{Painters} dataset and use it to generate images of size $256\times 256$. The model can transfer the painting style of a specific painter to the other painters, e.g., transferring the images of \textit{Cezanne} to images of other three painters \textit{Monet, Ukiyoe} and \textit{Vangogh}. \rev{In Fig.~\ref{pr_r}, we also compare our model with other reference models when given the same test image.}

\begin{figure}
\includegraphics[height=2.5in, width=2.5in]{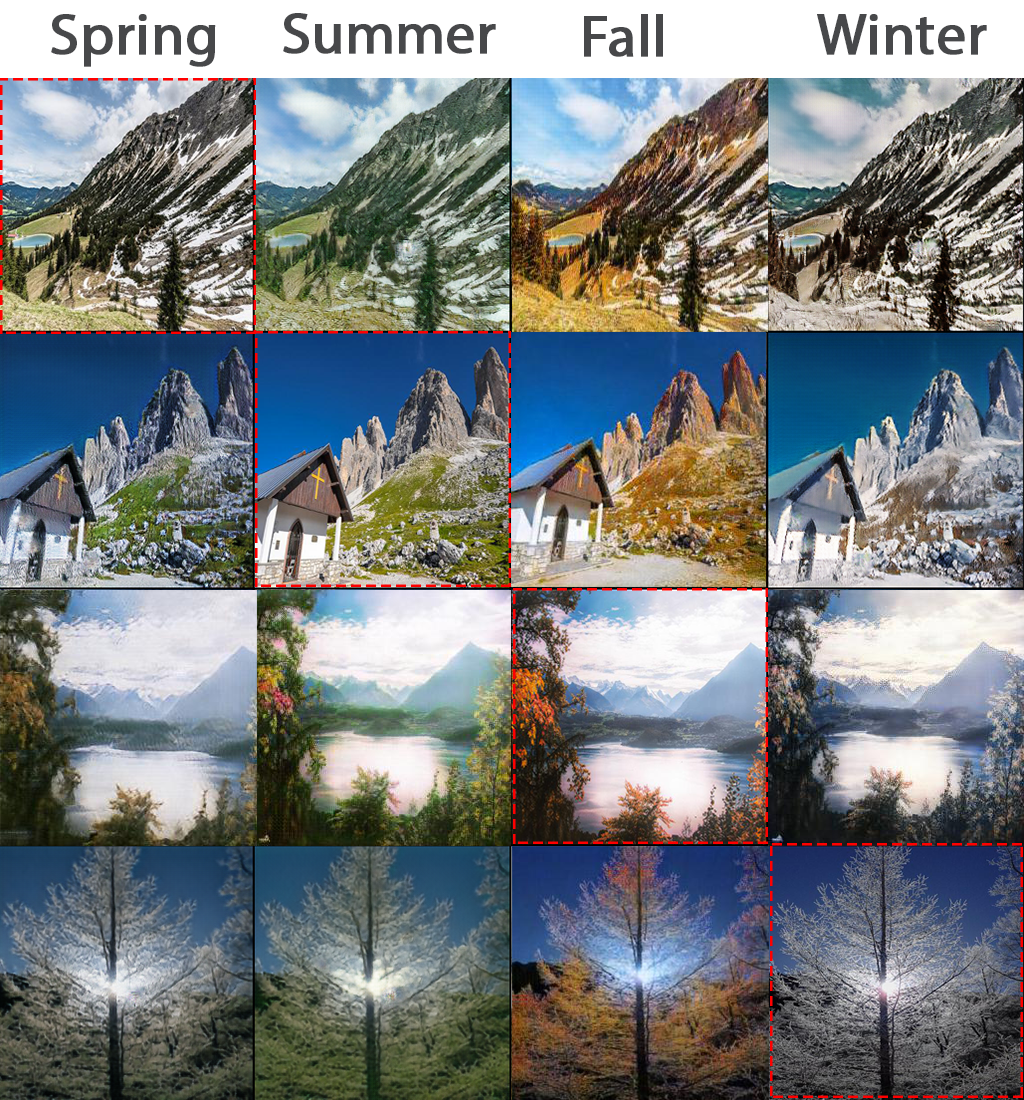}
\caption{\textit{Alps Seasons} translation results. The original images are displayed with a dashed square around. The other images are generated according to different seasons.}
\label{ar}
\end{figure}

\begin{figure}
\includegraphics[height=3.2in, width=3.2in]{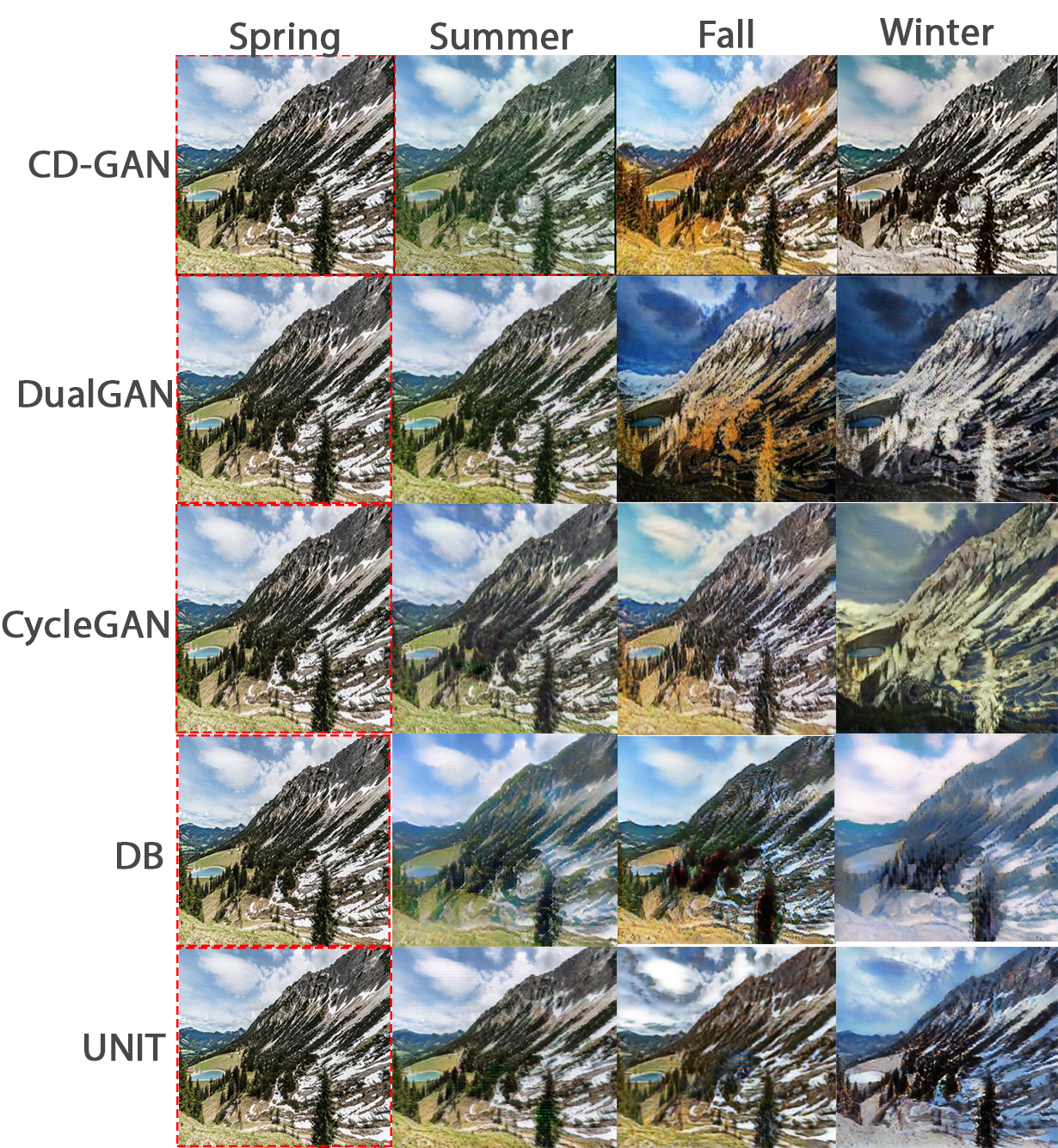}
\caption{\textit{Alps Seasons} translation results. The original images are displayed with a dashed square around. The other images are generated according to different seasons.}
\label{al_r}
\end{figure}

\textbf{Season transfer (Fig.~\ref{ar})}
The model is trained on the \textit{Alps Seasons} dataset. We use the trained model to generate images of different seasons. For example, we generate an image of summer from an image of spring and vice versa. \rev{In Fig.~\ref{al_r}, we also compare our model with other reference models when given the same test image.}

\begin{figure}
\includegraphics[height=3.2in, width=3.2in]{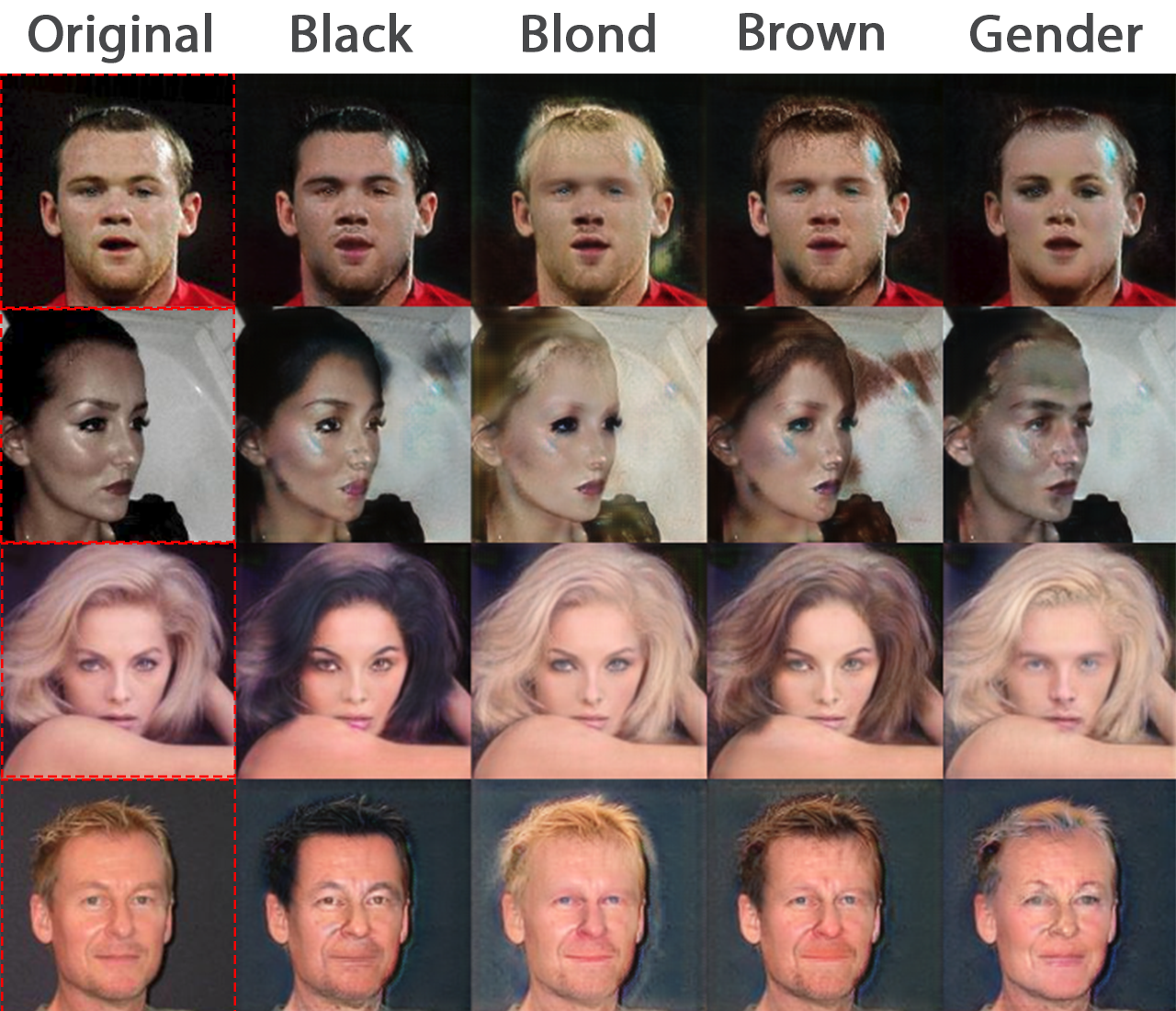}
\caption{Attribute-base face translation results. The original images are displayed with a dashed square around. The other images are generated according to different face attributes.}
\label{fr}
\end{figure}

\textbf{Attribute-base face translation (Fig.~\ref{fr})}
We train the model on \textit{CelebA} dataset for attribute-based face translation tasks. We choose 4 attributes, \textit{black hair}, \textit{blond hair}, \textit{brown hair}, and \textit{gender}. We then use our model to generate images with these attributes. For example, we transfer an image with a man wearing black hair to a man with blond hair, or transfer a man to a woman.

\section{Conclusion}
\label{conclusion}

In this paper, we propose a Cross-Domain Generative Adversarial Networks (CD-GAN), a novel and scalable model to conduct unsupervised multi-domain image-to-image translation. We show its capability of translating images from one domain to many other domain using several datasets. It still has some limitations. First, training could be unstable due to the training problem of GAN model. Second, the diversity of the generated images are constrained by the cycle consistency loss. We plan to address these two problems in the future work. 

\bibliographystyle{ACM-Reference-Format}
\bibliography{sample-bibliography}

\end{document}